\documentclass{article}

\usepackage[final]{nips_2019}

\usepackage[utf8]{inputenc} 
\usepackage[T1]{fontenc}    
\usepackage{hyperref}       
\usepackage{url}            
\usepackage{booktabs}       
\usepackage{amsfonts}       
\usepackage{nicefrac}       
\usepackage{microtype}      
\usepackage{multirow}
\usepackage{color}
\usepackage{amsmath,amssymb} 
\usepackage{pgfplots}
\pgfplotsset{compat=newest}
\usepackage{pdflscape}
\usepackage{graphicx}
\usepackage[export]{adjustbox}
\usepackage{algorithm}
\usepackage{mathtools}
\usepackage{algorithmic}
\usepackage{wrapfig,booktabs}

\usepackage{array}

\usepackage{wrapfig}

\title{ExpertMatcher: Automating ML Model Selection for Clients using Hidden Representations}

\author{
  Vivek Sharma \\
  MIT\\
  \texttt{vvsharma@mit.edu} \\
  \And
  Praneeth Vepakomma \\
  MIT \\
  \texttt{vepakom@mit.edu} \\
  \And
  Tristan Swedish \\
  MIT \\
  \texttt{tswedish@mit.edu} \\
  \AND
  Ken Chang \\
  MIT \\
  \texttt{kenchang@mit.edu} \\
  \And
  Jayashree Kalpathy-Cramer \\
  MGH/Harvard Medical School \\
  \texttt{kalpathy@nmr.mgh.harvard.edu} \\
  \And 
  Ramesh Raskar \\
  MIT \\
  \texttt{raskar@mit.edu} \\
}

\begin{document}

\maketitle

\begin{abstract}
Recently, there has been the development of Split Learning, a framework for distributed computation where model components are split between the client and server~\citep{vepakomma2018split}. As Split Learning scales to include many different model components, there needs to be a method of matching client-side model components with the best server-side model components. A solution to this problem was introduced in the ExpertMatcher~\citep{em} framework, which uses autoencoders to match raw data to models. In this work we propose an extension of ExpertMatcher, where matching can be performed without the need to share the client's raw data representation. The technique is applicable to situations where there are local clients and centralized expert ML models, but sharing of raw data is constrained.
\end{abstract}

\section{Introduction}

The Expert matching problem as described by \cite{em} is to assign input data from a client to the most appropriate expert model without any information other than the input data itself. In essence, the goal is to assign input data based on its likelihood of being drawn from the distribution of the expert model training data. We are interested in assigning an expert model to a given sample such that an improved performance is achieved. The expert matching problem is well-studied:~\citep{jacobs,jacobs1995methods,jacobs1997bias}, starting from the seminal work of~\cite{jacobs}, where the authors trained a versatile blend of experts for speaker vowel recognition and utilized a gating system to determine which of the systems should be utilized for each sample. Other lines of work try to avoid using a gating function~\citep{hinton,ahmed} by training one oracle model. \cite{aljundi} learns a gating function to make expert network assignments. More recently, with the ongoing advances in machine learning and big data, \cite{em} proposed a landscape for the expert matching problem and shows that by using an Autoencoder~(AE), it becomes possible to assign the most relevant expert model(s) given a sample data representation in a client-server setting. However, in their setting the client shares the raw data with the server. 

For many reasons raw data cannot always be shared. First, there is oftentimes a need to protect the privacy of the individual from whom the data originated (such as patient health data). Further, data is a valuable resource and many clients prefer not to have their data open to the public. Lastly, if data is shared, there will be additional storage requirements which may be cumbersome and expensive. Motivated by this observation, our approach does not require sharing of raw data directly. Instead, independent autoencoders are trained on both the server and client sides and only the intermediate hidden representation of the data is shared for expert network(s) assignment. 

We evaluate our proposed method on six datasets, such as objects, text, digits, biological and sensor, namely STl-10~\citep{stl}, MNIST~\citep{mnist}, HAR~\citep{har}, Reuters~\citep{reuters}, Non Line of Sight~\citep{nlos} and Diabetic Retinopathy~\citep{db}.  We are inspired from~\cite{em}, and we experimentally show that we can achieve similar performance as~\cite{em} to match the task (coarse-Level) assignment and reasonably good performance for class (fine-grained) assignment in a Split Learning distributed computation framework~\citep{vepakomma2018split,vepakomma2019reducing,vepakomma2018no}.

\begin{figure*}[t]
\centering
\vspace{-8mm}
{\includegraphics[width=0.8\columnwidth]{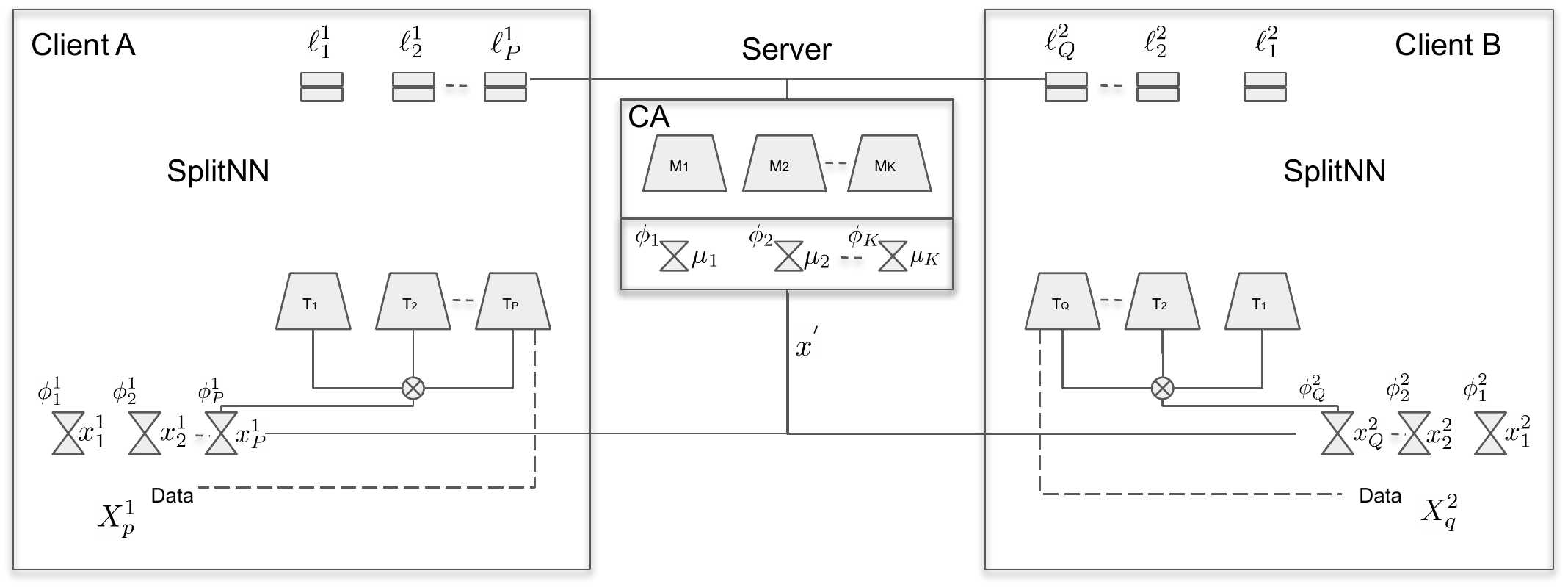}} 
\vspace{-4mm}
\caption{Pipeline of ExpertMatcher with hidden representations at the client and server for automatic model selection from a repository of server models based on their relevance to query data set hosted by the client. The matching is done based on encoded intermediate representations.}
\label{fig:pipeline}
\vspace{-5mm}
\end{figure*}

\begin{wrapfigure}{r}{0.5\textwidth}
\vspace{-8mm}
\centering
{\includegraphics[width=0.5\columnwidth]{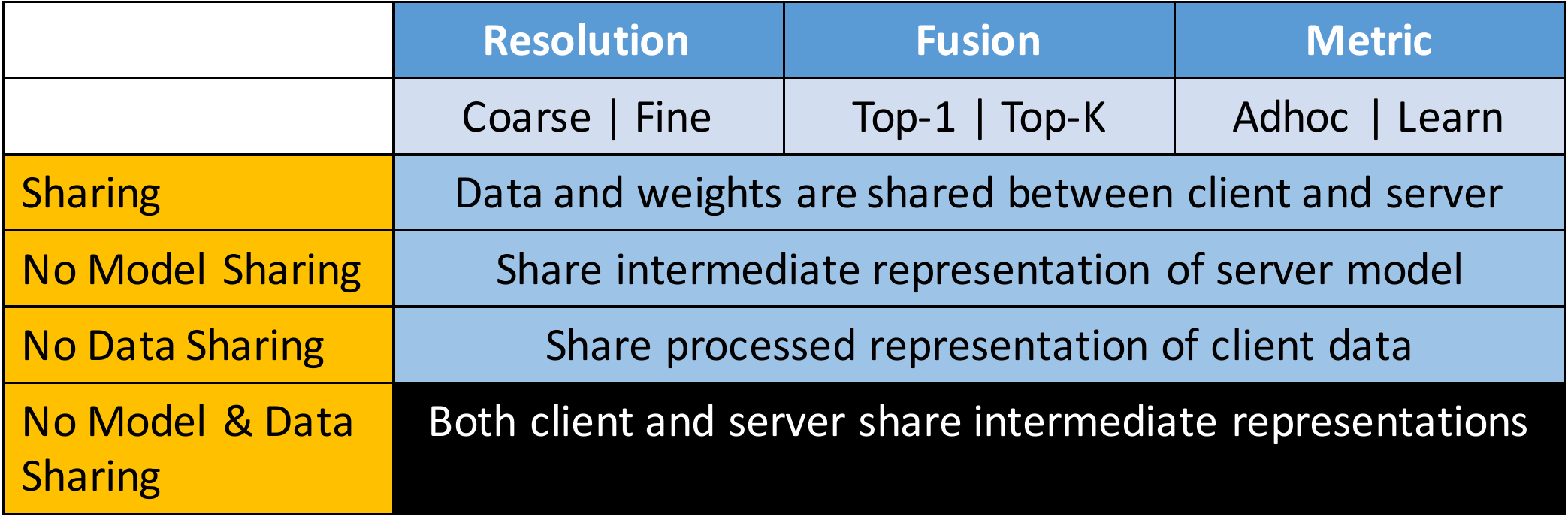}} 
\vspace{-6mm}
\caption{Landscape of ExpertMatcher, Source:~\cite{em}. This paper is relevant to the fourth setup where raw data or model is not shared between client and server, but rather some intermediate encoded representation is shared.}
\label{fig:landscape}
\vspace{-5mm}
\end{wrapfigure}

\section{Hidden Representation based ExpertMatcher} \label{sec:method}

The expert matching problem aims to assign a given sample the best model at hand for a given task. Figure~\ref{fig:landscape} shows the landscape of the ExpertMatching problem proposed by~\cite{em}. 
In this work, we consider the no sharing scenario (row fourth in Figure~\ref{fig:landscape}), where the client and server do not share the data and model, but just the low-dimensional intermediate representation. Furthermore, within Figure~\ref{fig:landscape} we consider (1) Resolution: coarse and fine level assignment; (2) Fusion: top-1; and (3) Metric: adhoc ( i.e. cosine similarity).

\begin{wrapfigure}{r}{0.5\textwidth}
\vspace{-9mm}
{\includegraphics[width=0.45\textwidth]{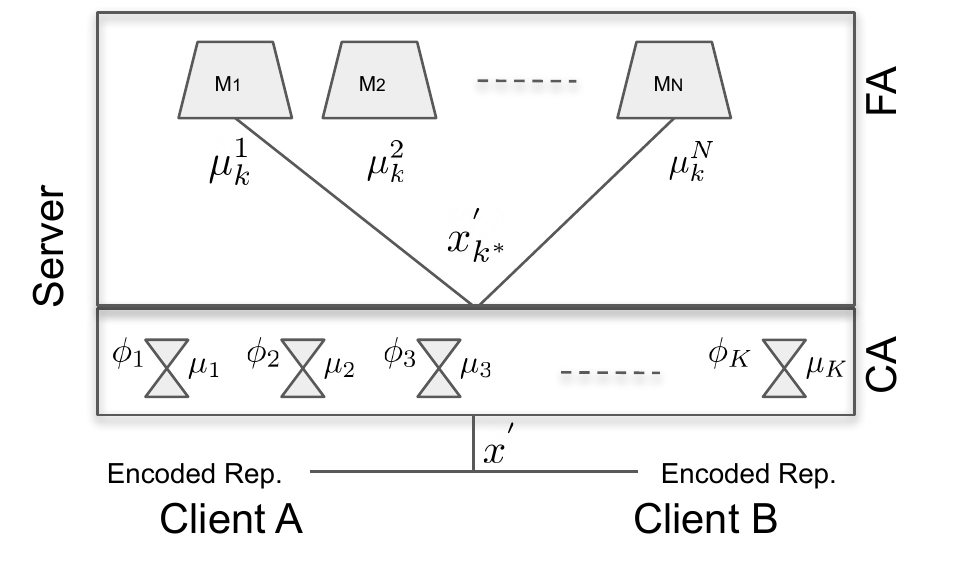}} 
\vspace{-5mm}
\caption{Overview of the coarse level dataset assignment~(CA) and fine-level model assignment~(FA) of the client's data in a hierarchical fashion. In contrast to~\cite{em}, for both CA and FA we use maximum cosine-similarity as a metric for model assignment. Further, the client shares the encoder representation rather than the raw samples~\citep{em}.}
\label{fig:ca_fa}
\vspace{-6mm}
\end{wrapfigure}

Now, we describe our proposed method to solve the ExpertMatcher problem without sharing raw data. In~\cite{em}, the raw data is shared with the Server. In our proposed method, the Client shares only a encoded hidden representation for both Coarse-level~(CA) and Fine-grained~(FA) expert matching. Figure~\ref{fig:pipeline} and Figure~\ref{fig:ca_fa} sketches the proposed method.

\textbf{Implementation:} We assume that we have $K$ pre-trained expert networks on the centralized server, each of these networks has its corresponding pre-trained autoencoders~(AE) $\phi_{K}$ trained on a task-specific dataset. Given the dataset on which the AE was trained on, we extract the hidden representations of the whole dataset and compute an average representation of the dataset $\mu_{k} \in \mathbb{R}^{d}, k \in \{1,\ldots,K\}$, where $d$ is the feature dimension. Assuming the dataset consist of $N$ object classes, we also compute the average representation of each class in the dataset $\mu_{k}^{n} \in \mathbb{R}^{d}, n \in \{1,\ldots,N\}, k \in \{1,\ldots,K\}$. 

The clients (Client A and Client B) utilize a similar approach as the server, where the clients train their unique AE's one for each $p^{th}$ and $q^{th}$ datasets, Client A: $p \in \{1,\ldots,P\}$ and Client B: $q \in \{1,\ldots,Q\}$. Lets assume for Client A, the intermediate features extracted from a hidden layer for a sample $X_{p}^{1}$ is given as $x_{p}^{1}=\phi_{p}^{1}(X_{p}^{1})$, similarly for Client B, $x_{q}^{2}=\phi_{q}^{2}(X_{q}^{2})$. For brevity, we denote the intermediate representation coming from any client as $x^{'}$.

For coarse assignment~(CA) of clients data: hidden representation $x^{'}$, we assign an server AE, $k^{\ast} \in \{1,\ldots,K\}$ which has maximum cosine similarity of $x^{'}$ with $\mu_{k}$. 

For fine-grained~(FA)  of clients data: hidden representation $x^{'}$, we assign an expert network $M_{n}, n \in \{1,\ldots,N\}$ that has the maximum cosine similarity of $x_{k^{\ast}}^{'}$ with $\mu_{k}^{n}$.

Finally after the assignment of the given sample to the model, one can easily train a SplitNN type architecture~\cite{vepakomma2018split,vepakomma2019reducing,vepakomma2018no}.

In the current setup, a weak level of privacy is preserved as the server does not have access to the raw client's data, but rather a very low dimensional encoded representation.

Note that there is a shortcoming of this approach. If the server has no AE model dedicated to the client data, the wrong assignment of client data takes place because of maximum cosine similarity criteria - this can be resolved by adding an additional model on the server that performs a binary classification: if the client data matches the server data or not. 

\section{Experiments} \label{sec:experiments}
We present our evaluation of several different datasets. We first describe the dataset, metric, implementation details, followed by a thorough analysis of the proposed methods and comparison to~\cite{em} for both coarse assignment~(CA) and fine-grained assignment~(FA).

\textbf{Datasets and Implementation Details.} 
We demonstrate our method on challenging datasets which
cover domains such as objects, text, digits, biological and raw sensors. The datasets are summarized in Table~\ref{table:stats}. For a fair comparison with~\cite{em}, we use the same splits for Server, client A and Client B. 

For all the image data types, we resize them to $28 \times 28$, and then flattening it to 784 dimensions. While for other datasets, we use 1D adaptive average pooling (\texttt{AdaptiveAvgPool1d}) to transform the input data to 784 dimensions. \textbf{AE:} All the AEs use a single-layer MLP encoder-decoder ($\mathbb{R}^{784} \rightarrow \mathbb{R}^{128} \rightarrow \mathbb{R}^{784}$). We use Adam optimizer for model training. We train the AE models for 45 epochs with a learning rate of $10^{-2}$, which is then manually decreased by a factor of 10 every 15 epochs.

Note that, in this work, the server shares the pseudorandom number generator seeds with the clients to make the same random initialization for PyTorch model training (e.g. \texttt{torch.manual\_seed(0)}). In future work, we plan to study the impact of random initialisation when the seed is not shared between server and clients.

\begin{table*}[ht]
\small
\tabcolsep=1.5mm
\begin{center}
\caption{\textbf{Datasets}. ``\#S'' denotes the number of samples, ``\#C'' denotes the true number classes, and LC/SC is the class balance largest class~(LC) to smallest class~(SC). } 
\label{table:stats}
\resizebox{10cm}{!}{
\begin{tabular}{l|cccccc|c}
\toprule
&	 STL-10 & MNIST & HAR &  REUTERS & NLOS & DB& ALL  \\
\cmidrule{1-7}
Type & Object & Digits & Sensors & Text & Sensor & Biological & \\
\#C & 10 & 10 & 6 & 4 & 3 & 3 & \\
\#S & 13k &10k & 10299 &10k &45096  & 3540 &\\
Dim. & 32px & 28px & 561 & 2000 & $640\times480$ & 512px \\
LC/SC~(\%) & 10/10 &  11.35/8.92 & 19/14 & 43.12/8.14 &  33.33/33.33& 33.33/33.33&\\
\midrule
Server  &   6500&   5000&   5151&   5000& 22548 & 1770    \\ 
Client A &   3250&   2500&   2574&   2500& 11274 & 885 & 22983\\
Client B &   3250&   2500&   2574&   2500& 11274 & 885 & 22983\\
\bottomrule
\end{tabular}}
\vspace{-5mm}
\end{center}
\end{table*}

\textbf{Evaluation Metric.} For both CA and FA, we use maximum cosine similarity for making a model assignment~(CA) / class assignment~(FA) and then computing the accuracy between predicted class and target class.

\textbf{Multiple Clients.} In the current work, we consider two clients (Client A and Client B) for evaluation. However, the amount data available to each client is the same. In future work we plan to study the impact of the varying proportion of samples assigned to the clients for model training and assignment. Our idea is easily scalable to many more clients, and is not limited to the two client scenario presented here.

\subsection{Coarse-level dataset assignment~(CA)}

In Table~\ref{table:ca}, we compare the results for the coarse level dataset assignment with~\cite{em}. In case of~\cite{em}, the authors use minimum reconstruction error (i.e. MSE) as the criteria for making the sample assignment when the client shares the raw data with the server. 

These results suggest that autoencoders trained on non-overlapping datasets at the client or server side perform similarly to \cite{em} - where the client shares the raw data with the server. Note we are not sharing our raw data - but only hidden representation. Further, we note that autoencoders are effective in learning the underlying representation of the dataset even if the client and server AE's are trained on different datasets. This is a surprising result, as training with non-overlapping data seems to lead to representations that are close via cosine similarity. In this work, we ensure models are initialized with the same seed, which encourages the models to converge to similar representations as long as the training samples are coming from the same underlying distribution. Otherwise, we expect the order of learned filters would not be preserved and cosine similarity would fail. We note that both Client A and Client B are 99\% accurately assigned to their corresponding autoencoders.

Note that in the financial services industry, we have a similar situation where we are interested in finding an expert method for handling our specific problem at hand without loss of confidential information. A similar example dataset we have our experimental setup is REUTERS that contains newswire articles, which may be similar to financial analysts reports or other valuable text-based data that may not want to be shared with third parties. We think our approach to finding an expert model without sharing the raw data can easily be deployed to the financial industry where confidential information may not want to be shared in raw form. Sharing encoded representations provides some weak privacy, but also has other benefits such as lower bandwidth requirements leading to decreased latency and more efficient storage of data being shared between client and server.

\begin{table}[ht]
  \caption{Coarse-level dataset assignment using cosine-similarity as the assignment metric for computing accuracy~(\%). Note that~\cite{em} use MSE loss as the assignment metric.}
  \label{table:ca}
  \centering
  \vspace{-1mm}
  \resizebox{10cm}{!}{
  \begin{tabular}{l|cccccc|c}
    \toprule
         & MNIST &  STL-10  & HAR & REUTERS & NLOS & DB & Average   \\
    \midrule
    Client A~\citep{em}     & 100.0 & 100.0  & 100.0  & 99.64  & 99.92  & 96.49 & 99.34 \\
    Client A~(\textbf{ours})   & 99.36 & 99.93 & 99.80 & 96.36 & 99.59 & 100.0 & 99.30    \\
    \midrule
    Client B~\citep{em}     & 100.0 & 100.0  & 100.0  & 99.56 & 99.89  & 95.36 & 99.13    \\
    Client B~(\textbf{ours})   & 99.80 & 99.96 & 99.80 & 96.40 & 98.28 & 100.0 & 98.72    \\
    \bottomrule
  \end{tabular}}
\end{table}

\subsection{Fine-grained class assignment~(FA)}
In Table~\ref{table:fa}, we compare the results for the fine-grained class assignment with~\cite{em}. It is interesting to observe that the autoencoder trained on the client-side and the server-side seems to perform well even if they are trained independently with different samples although coming from the same underlying distribution - this shows that autoencoders are effective at learning the class identity representation. A drop in performance for our method is expected, as the client's AE model has access to  much smaller set of examples than the server.  


\begin{wraptable}{r}{5.5cm}
\small
\tabcolsep=1mm
\begin{center}
\vspace{-18mm}
  \caption{Fine-grained class assignment using cosine-similarity as the assignment metric for computing accuracy~(\%).}
  \label{table:fa}
  \vspace{-1mm}
  \resizebox{7cm}{!}{
  \begin{tabular}{lccc}
    \toprule
    Dataset     & \#C & Client A     & Client B \\
    \midrule
    MNIST~\citep{em}   &10 &  84.36 & 83.40  \\
    MNIST~(\textbf{ours}) &10 &  71.4 & 71.64  \\
    \midrule
    NLOS~\citep{em}   &3  &  71.78 & 71.26 \\
    NLOS~(\textbf{ours})  &3  &  59.40 & 52.40 \\
    \midrule
    DB~\citep{em}      &3  &  41.47 & 44.41 \\
    DB~(\textbf{ours})  &3  &  44.30 & 53.33 \\
    \bottomrule
\end{tabular}}
\end{center}
\vspace{-8mm}
\end{wraptable}

\section{Conclusion} \label{sec:conclusion}
In this work, we propose a novel ExpertMatcher based model selection where the raw data of remote clients are not shared with the central server. In contrast to the previous method, the server and client both train a model independently and the intermediate hidden representations are shared between them for either coarse or fine-grained matching respectively. We believe our approach to finding an expert model via hidden representations provides a path for a new paradigm of distributed machine learning architectures where sharing of raw data is not feasible.

\textbf{Acknowledgements}
V. Sharma would like to thank Karlsruhe House of Young Scientists~(KHYS) for funding his research stay at MIT.

\bibliographystyle{authordate1} 
\bibliography{main}

\end{document}